\pdfoutput=1

\documentclass[11pt]{article}

\usepackage[final]{acl}

\usepackage{times}
\usepackage{latexsym}
\usepackage{amssymb}
\usepackage{xcolor}
\usepackage{amsmath}
\usepackage{multirow}
\usepackage{graphicx}
\usepackage{paralist}
\usepackage{booktabs}
\usepackage{array}
\usepackage{multicol}
\usepackage{newfloat}
\usepackage{listings}

\usepackage[T1]{fontenc}

\usepackage[utf8]{inputenc}

\usepackage{microtype}

\title{{W}hat do {L}arge {L}anguage {M}odels {L}earn about {S}cripts?
}

\author{Abhilasha Sancheti \\
  University of Maryland, College Park \\
  Adobe Research\\
  \texttt{sancheti@\{umd.edu,adobe.com\}} \\\And
  Rachel Rudinger \\
  University of Maryland, College Park \\
  \texttt{rudinger@umd.edu} \\}

\begin{document}
\maketitle
\begin{abstract}
Script Knowledge~\citep{schank1975scripts} has long been recognized as crucial for language understanding as it can help in filling in unstated information in a narrative. However, such knowledge is expensive to produce manually
and difficult to induce from text due to reporting bias~\citep{gordon2013reporting}. In this work, we are interested in the scientific question of whether explicit script knowledge is present and accessible through pre-trained generative language models (LMs). To this end, we introduce the task of generating full event sequence descriptions (ESDs) given a scenario as a natural language prompt. Through zero-shot probing, we find that generative LMs produce poor ESDs with mostly omitted, irrelevant, repeated or misordered events. To address this, we propose a pipeline-based script induction framework (\texttt{SIF}) which can generate good quality ESDs for unseen scenarios (e.g., bake a cake). 
\texttt{SIF} is a two-staged framework that fine-tunes LM on a small set of ESD examples in the first stage. In the second stage, ESD generated for an unseen scenario is post-processed using RoBERTa-based models to filter irrelevant events, remove repetitions, and reorder the temporally misordered events. 
Through automatic and manual evaluations, we demonstrate that \texttt{SIF} yields substantial improvements ($1$-$3$ BLEU points) over a fine-tuned LM. However, manual analysis shows that there is great room for improvement, offering a new research direction for inducing script knowledge\footnote{Code and dataset are available at \url{https://github.com/abhilashasancheti/script-generation}}.
\end{abstract}

\section{Introduction}
Scripts are structured commonsense knowledge in the form of event sequences that characterize commonplace scenarios, such as, eating at a restaurant~\citep{schank1975scripts}. Scripts are fundamental pieces of commonsense knowledge that humans share and assume to be tacitly understood by each other. When someone says ``I went to a restaurant for lunch", our script knowledge allows us to infer that a waiter would have taken the order, the speaker would have eaten the lunch, payed for it, and tipped the waiter, even if these events are not explicitly mentioned. Knowledge of scripts, whether implicit or explicit, has been recognized as important for language understanding tasks~\citep{miikkulainen1995script,mueller2004understanding}.

Earlier efforts to automatically induce scripts from text on a large scale include \citet{chambers2008unsupervised} who treat the problem of script induction as one of learning narrative chains using textual co-occurrence statistics.
\begin{figure}[t!]
    \centering
    \includegraphics[width=\columnwidth]{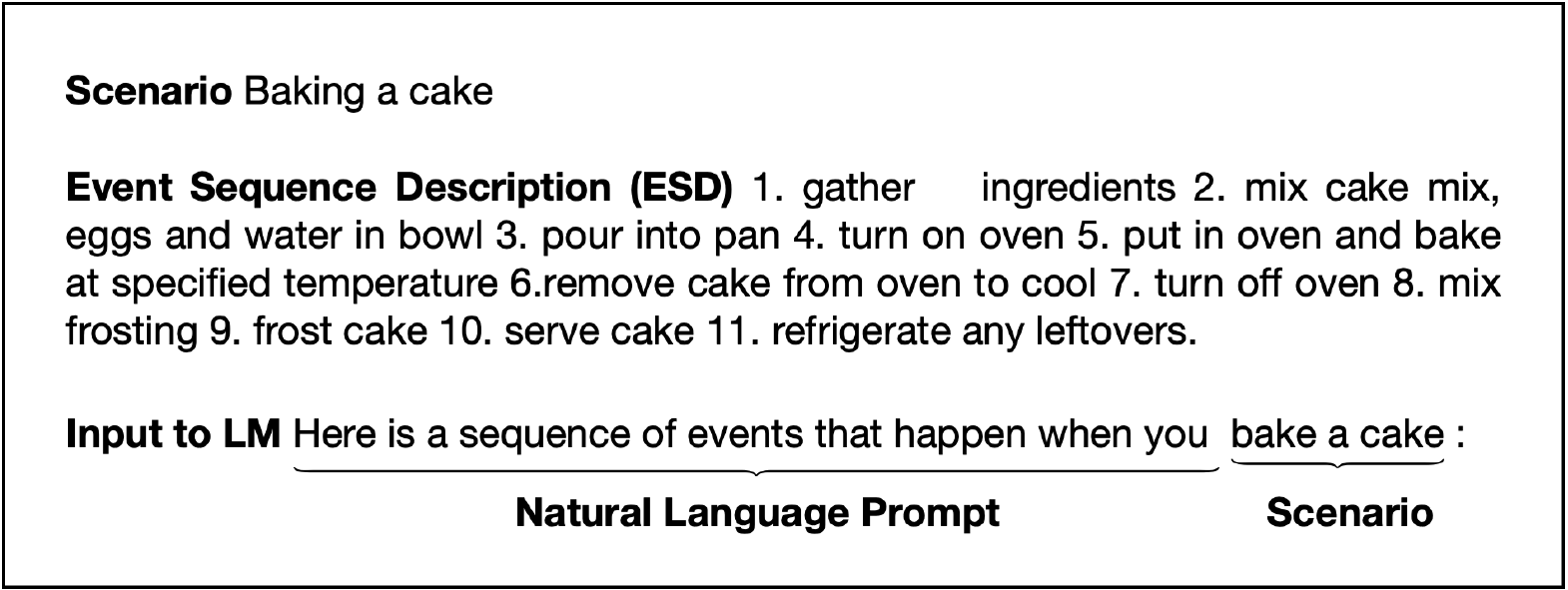}
    \caption{Sample event sequence description (ESD) from~\citet{wanzare2016crowdsourced} for \textsc{baking a cake} scenario. We use natural language prompts (Table~\ref{fig:probing-prompts}) to \textit{generate} \textit{completely ordered} ESDs for evaluating extent of script knowledge accessible through LMs.}
    \label{fig:task}
\end{figure}
However, reporting bias~\citep{gordon2013reporting} remains an obstacle for script induction as many events are not mentioned explicitly in text, relying on the reader's ability to infer missing script-related events.
Moreover, manual creation of such knowledge resources is challenging due to the wide coverage and complexity of relevant scenario knowledge. 
Although crowdsourced efforts~\cite{singh2002open,regneri-etal-2010-learning,modi2017inscript,wanzare2016crowdsourced,ostermann2018mcscript,ostermann2019mcscript2} address these issues and acquire script knowledge in the form of ESDs, the collected datasets are small, domain-specific, and crowdsourcing is not scalable.

With the success of pre-trained language models (henceforth,  PLMs)~\citep{devlin2018bert,liu2019roberta,radford2019language} in various natural language understanding tasks, we are interested in investigating the extent and accessibility of explicit script knowledge present in PLMs. In this work, unlike cloze-based script evaluations~\citep{chambers2008unsupervised,mostafazadeh2016corpus} which LMs are uniquely optimized for~\citep{rudinger-etal-2015-script}, we evaluate PLMs on the ability to \textit{fully generate} event sequence descriptions (ESDs)~\citep{regneri-etal-2010-learning} in free-form natural language (Figure~\ref{fig:task}). This is a challenging task as scripts are complex structures with varied granularity of describing a scenario (e.g., starting from going to grocery store to buy ingredients or starting with finding a recipe for \textsc{baking a cake} scenario), and the requirement to produce all the scenario-\textit{relevant} events in the correct temporal \textit{order}. 

To this end, we first probe LMs via carefully crafted prompts to analyze the quality of ESDs generated in a zero-shot setting (\S\ref{sec:probing}) and find that the generated ESDs are of poor quality with many scenario-irrelevant, repeated, temporally misordered, and missing events. To address this we propose a, LM-agnostic, pipeline-based script induction framework (\S\ref{sec:induction}), \texttt{SIF}, which can generate good quality ESDs for novel scenarios that LM has not seen during the training phase of the framework. \texttt{SIF} is a two-staged  framework with fine-tuning LM on a small set of ESDs as the first stage followed by a three-stepped post-processing stage which corrects the ESDs generated from a fine-tuned LM for irrelevant, repeated, and temporally misordered events. 
This work makes the following \textbf{contributions}: 
\begin{itemize}
    \item We present an analysis of the extent of script knowledge accessible through LMs using probing techniques, in a zero-shot setting, via the task of generating full ESDs from natural language prompts.
    \item We propose script induction framework that can generate ESDs for held-out and novel scenarios drawn from a different distribution.
    \item We present automatic and manual evaluation of the generated ESDs, establishing the viability of our framework and paving way for future research in this direction.
\end{itemize}
\section{Related Work}
\noindent \textbf{Narrative Chain Induction}~There has been a growing body of research into statistical script learning systems which can automatically infer implicit events from text. Seminal work by~\citep{chambers2008unsupervised,chambers2009unsupervised} describe
a number of simple event co-occurrence based
systems that infer (verb, dependency) pairs (known as narrative events)
with partial-ordering related to one or multiple
participants~\citep{pichotta2014statistical} in discourse (known as narrative chains). As statistical co-occurrences cannot capture long-range dependencies between events, \citet{pichotta2016learning} represent events using LSTM leading to improved narrative cloze task performance. 
However, much of the information about events is usually left implicit in text. Moreover, narrative events are highly abstracted~\citep{ostermann2020script} and cloze task is insufficient to evaluate script knowledge~\citep{chambers2017behind}. Therefore efforts have been made
to acquire crowdsourced ESDs~\citep{singh2002open,regneri-etal-2010-learning,modi2017inscript,wanzare2016crowdsourced,ostermann2018mcscript,ostermann2019mcscript2} and to learn similar events in a scenario using unsupervised~\citep{regneri-etal-2010-learning} and semi-supervised~\citep{wanzare2017inducing} approaches.

\noindent \textbf{Temporal Ordering and Relevance}~
Previous works~\citep{modi2014inducing,wanzare-etal-2017-inducing,lyu-zhang-wikihow:2020} have investigated induction or prediction of temporal ordering of prototypical events. Others have predicted next~\citep{pichotta2016using} or related~\citep{lyu-zhang-wikihow:2020} events in natural language form. \citet{zhou2019learning} acquire commonsense procedural knowledge directly from natural language source, like wikiHow, by learning representations for scenarios and events which are predictive of both relevance of event to the scenario and temporal ordering.
\citet{zhang2020analogous} propose a non-learning based approach to predict fixed-length events given an unseen scenario and related scenarios with their events. A recent work~\citep{sakaguchi2021proscript} generates partially-ordered scripts using PLMs by predicting events and edges for partial-order while we are interested in completely ordered event descriptions. \citet{lyu-etal-2021-goal} propose the task of goal oriented script construction for multilingual wikiHow dataset and propose generation and retrieval-based approaches. However, their generation-based approach using LM only involves fine-tuning. We focus on different LMs to evaluate them on the task of generating scripts both in zero-shot and fine-tuning settings. Our proposed framework is shown to outperform the fine-tuning approach. 

\noindent \textbf{Knowledge-acquisition from PLMs}~ With the success of PLMs~\citep{devlin2018bert,liu2019roberta,radford2019language} in various natural language understanding tasks, a number of works investigate how commonsense knowledge is captured in these models~\citep{feldman2019commonsense,petroni2020context,weir2020probing,shwartz2020unsupervised}. Successful efforts have been made to induce relational~\cite{bouraoui2020inducing}, numerical ~\citep{lin2020birds}, temporal~\citep{zhou2020temporal} and commonsense knowledge in PLMs using fine-tuning. 

Unlike prior works, we focus on investigating the extent and accessibility of explicit script knowledge from PLMs via probing techniques and inducing such knowledge in them using a pipeline-based framework to generate full ESDs for novel scenarios in free-form natural language.
\begin{table*}
\centering
\scriptsize
\resizebox{0.95\textwidth}{!}
{\begin{tabular}{c p{16cm}}
\toprule
$1$&\textbf{these are the things that happen when you bake a cake: }3 tablespoons (30ml) of unsalted butter, at room temperature 2 tablespoons (30ml) of granulated sugar 2 tablespoons (30ml) of brown sugar, sifted 1 large egg, at room temperature 1 teaspoon vanilla extract 3 large egg yolks 1 cup (350ml) buttermilk To the bowl of an electric mixer fitted with a paddle attachment, add the butter, sugar, brown sugar, egg yolks, and vanilla extract and beat until light and fluffy. Add the buttermilk, then continue beating until fully incorporated. Sift the flour, baking powder, and salt into the bowl of the mixer and beat\\
\hline 
$2$&\textbf{here is an ordered sequence of events that occur when you bake a cake: }1. The cake is baked. 2. The cake is cooled. 3. The cake is served at the table. 4. The dessert is placed on the table. 5. The cake is placed in a baking pan. 6. The cake is transferred to the oven. 7. The cake is put to the middle of the oven and baked. 8. The cake is left in the oven until it is done. 9. The cake is removed from the oven and placed on a plate. 10. The cake is transferred to a cake rack. 11. The cake is placed on a cake plate. \\
\hline
$3$&\textbf{describe baking a cake in small sequences of short sentences: }``I want my cake to be soft and moist." (And, you know, don't do this: ``If I use a pastry bag, I can get it back to me in less than two minutes.") You may be looking for the most simple recipe for cake, but baking requires skill and a lot of practice. You don't need to learn how to make a cake recipe. You don't even need to know the difference between the cake and the sponge. All you need is a great recipe and a couple of ovens and an oven thermometer. 2. Use a mixer. Mixers are really, really good. (A great idea is to invest in one of those mini, little, portable ones.\\
\hline
$4$&\textbf{here is a sequence of events that happen while baking a cake: 1. }The cake goes into the oven. 2. The cake cools down in the oven, because the water vapor from the oven condenses onto the top of the cake. 3. The cake is cooled and the frosting set (the sugar crystals are not allowed to fall off). 4. The frosting is set (a little more water vapor condenses onto the top of the frosting). 5. The frosting is set (more water vapor condenses onto the top of the frosting). 6. The cake cools down (the water vapor cools off). 7. The cake comes out of the oven and it is ready to eat. How is this cake not a cake? The first step is that the cake is a cake.\\
\hline
$5$&\textbf{here is an ordered sequence of events that occur when you bake a cake: 1. }You have a cake. 2. You have a batch of flour (or other ingredient, such as water, sugar or corn syrup) 3. You have your flour prepared 4. You have your flour mixture 5. You have your flour mixture pre-floured 6. You have your batter prepared 7. You have your batter made 8. You have your batter placed on a baking sheet 9. You have your batter removed from the oven 10. You have your cake cooled in the tray (you must place your cake in the tray before you take it out of the oven) 11. You take your cake out of the tray and remove it from the oven 12. You eat your cake, or leave it out to cool.\\
\hline
$6$&\textbf{here is a sequence of events that happen while baking a cake: 1. get a cake mix }2. pour the cake mix into the pan 3. let it bake for 10 minutes 4. get out the oven lid and turn it on 5. let it bake for another 10 minutes 6. get out the oven lid and flip it on 7. let it bake for another 10 minutes 8. get out the oven lid and switch it back on 9. let it bake for another 10 minutes 10. get out the oven lid, turn it back on, and let it bake again 11. get out the oven lid and get out the knife 12. bake the cake again, and then let it cool on a rack 13. use the cake mix to make cupcakes 14. cut the cake, and then use the cupcake for serving!\\
\hline
$7$&\textbf{these are the things that happen when you bake a cake: 1. get a cake mix 2. gather together other ingredients }that you'll need for your cake (coconut oil, powdered sugar, eggs, etc.) 3. preheat oven to 350 4. pour on flour and mix by hand 5. pour in egg and mix by hand 6. add in powdered sugar and mix by hand 7. add in coconut oil and mix by hand 8. let it all sit for a minute 9. add in your flour and mix by hand 10. add in your sugar and mix by hand 11. add in your flour and mix by hand 12. add in your flour and mix by hand 13. add in your flour and mix by hand 14. add in your flour and mix by hand 15. add in your egg and mix by hand 16. add in your salt\\
\hline
$8$&\textbf{describe baking a cake in small sequences of short sentences: 1. get a cake mix 2. gather together other ingredients }3. start to bake 4. bake a cake 5. take it to the oven 6. the cake is done 7. is it done? 8. is it still hot? 9. what is that crust? 10. what does the cake look like? If we think of a sequence of a few hundred or so words, we can take the basic idea of this example and say that we could write an equivalent, yet shorter, sentence: 1. Get a cake mix 2. Get together the ingredients 3. Bake a cake 4. Take it to the oven 5. The cake is done 6. Is it done? 7. is it hot? 8. what is that crust? 9. what does the cake look like\\
\bottomrule
\end{tabular}}
\caption{\label{tab:probing} Scripts generated from GPT$2$-L for \textsc{baking a cake} scenario with \textbf{bold-faced} prompts.}
\end{table*}
\begin{figure}[t!]
    \centering
    \includegraphics[width=0.45\textwidth,height=3.0cm]{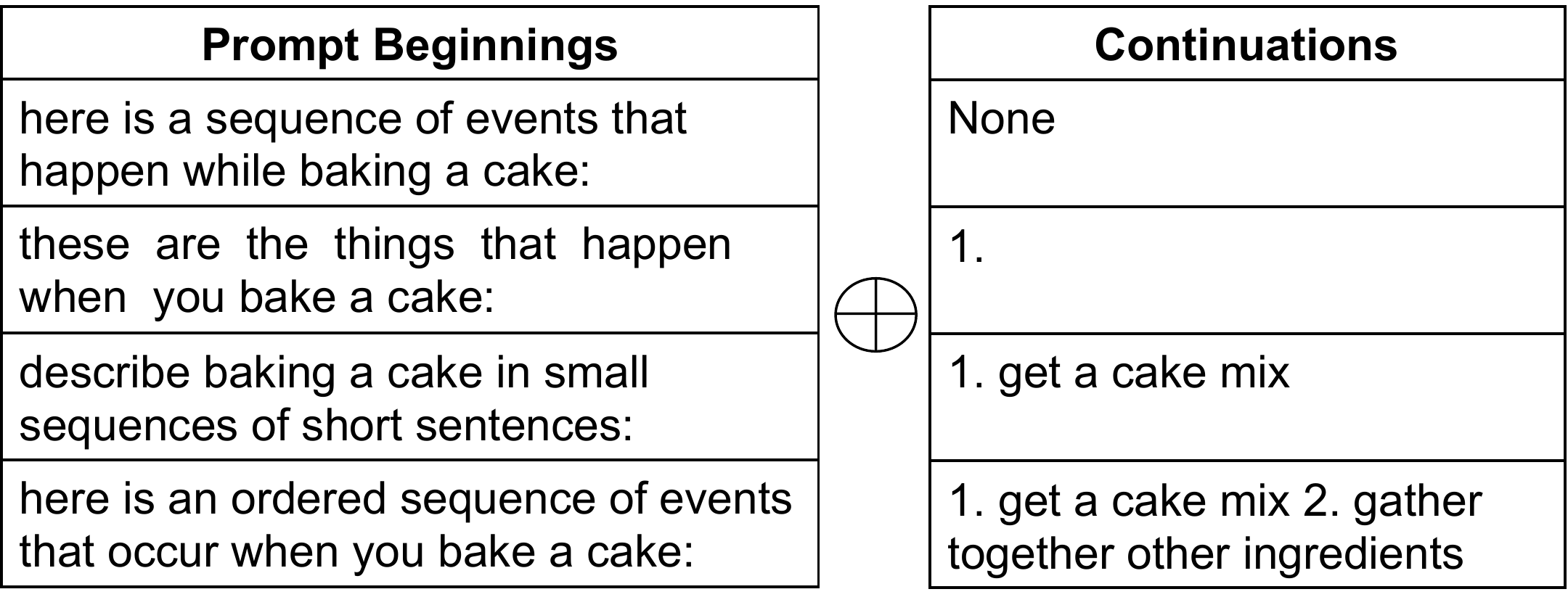}
    \caption{Different prompt formulations for \textsc{baking a cake} scenario for probing. $16$ prompts are created by combining a prompt beginning with a continuation. }
    \label{fig:probing-prompts}
\end{figure}
\section{Probing for Script Knowledge} \label{sec:probing}
We design a zero-shot probing experiment to evaluate PLMs' ability to generate ESDs by carefully selecting natural language prompts, which LMs are known to be sensitive to~\citep{bouraoui2020inducing}.
We experiment with $16$ manually crafted prompts\footnote{We also experiment with capitalized prompts but did not find significant change in the quality of generations.} (Table~\ref{tab:prompts}) with different phrasing and levels of conditioning to enquire large versions of GPT$2$, BART, and T$5$ for script knowledge. The intuition behind these prompts is similar to asking questions (prompts) to a knowledge source in various ways to get the required answer (ESD for a scenario).

BART and T$5$ were not able to output anything except the input prompt or start, end, and pad tokens and hence we only present qualitative outputs from GPT$2$, when probed with various prompts for \textsc{baking a cake} scenario, in Table~\ref{tab:prompts}. 
We observe that the quality of generated ESDs vary for different prompts. Although GPT$2$ is able to generate some scenario-relevant events with just the prompt beginnings and no continuations (e.g., $1$ and $2$ in Table~\ref{tab:prompts}), the ESDs are incomplete with many auxiliary details, and incorrect event ordering (e.g., `3. The cake is served at the table' before `6. The cake is transferred to the oven.' in $2$). It sometimes outputs (e.g., $4$) narrations rather than procedural descriptions. As generation from scratch is an open-ended task, we use a prompt with a number to guide GPT$2$ to generate a procedural script. Although $4$ and $5$ are more procedural, the events are still at a coarse-grained level with most of the intermediate events missing. To further guide the generation towards a fine-grained level, we condition the generation on a few events (manually curated by authors looking at sample ESDs) along with the prompt beginning. This helps us in examining whether GPT$2$ has temporal knowledge about the events related to a scenario. Conditioning on the events results in a better quality ESD (e.g., $6,7,8$).  However, there is a repetition of events (`let it cool for another 10 minutes’ in $6$, ‘add in your flour and mix by hand’ in $7$) in addition to wrong event ordering, irrelevant (e.g., `is it hot?' in $8$) and missing events.
As GPT$2$ produces poor quality ESDs in this zero-shot setting with BART and T$5$ not even being able to output any events, we propose a script induction framework detailed in the following section.
\begin{figure}[t!]
    \centering
    \includegraphics[width=\columnwidth,height=3.3cm]{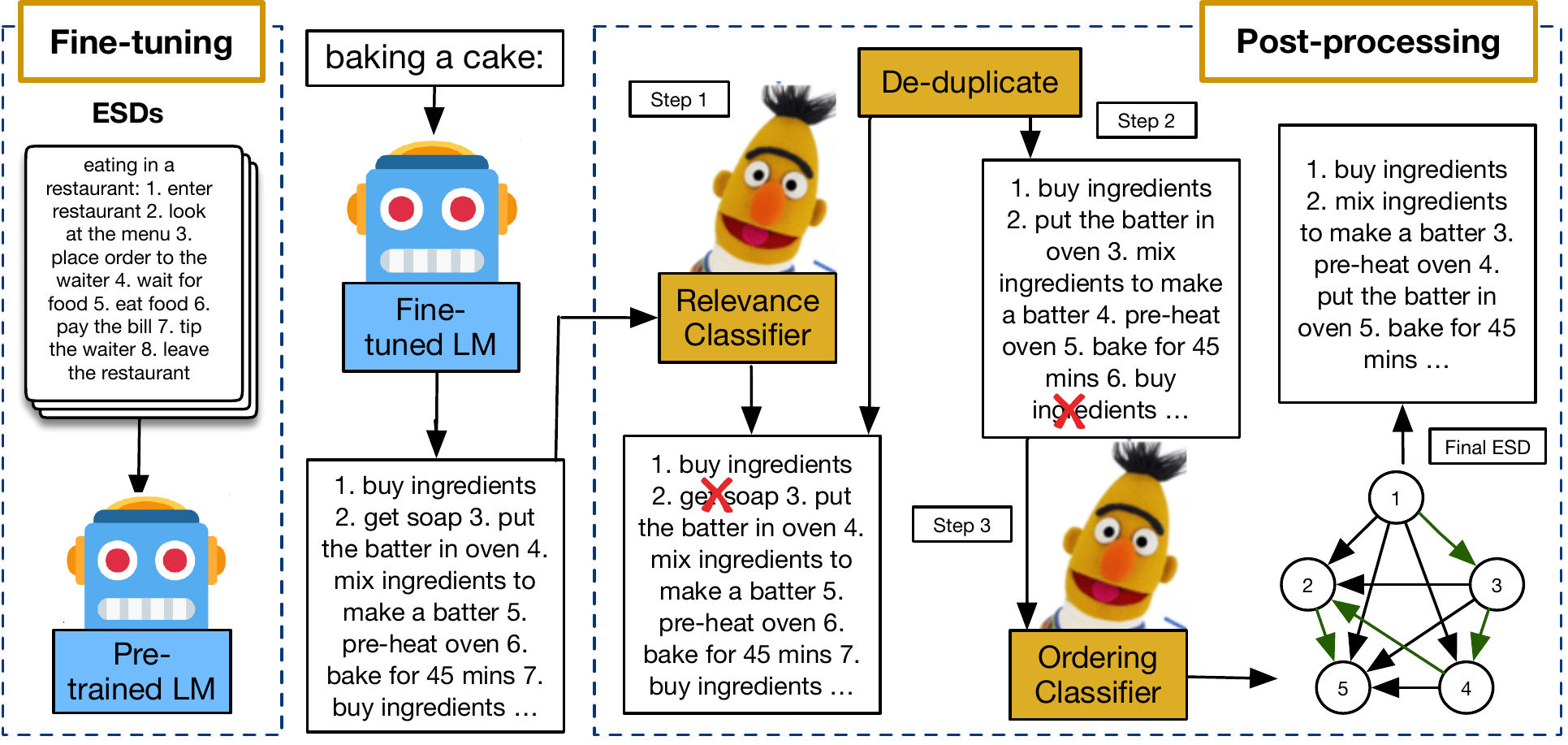}
    \caption{\texttt{SIF}: Pre-trained LM is fine-tuned on DeScript~\citep{wanzare2016crowdsourced}. Generated scripts are then post-processed with RoBERTa-based classifiers to correct for event relevance (Step 1), repetition (Step 2), and temporal ordering (Step 3).}
    \label{fig:framework}
\end{figure}
\section{\texttt{SIF:} Script Induction Framework} \label{sec:induction}
In this section, we provide details on our pipeline-based script induction framework, \texttt{SIF} (Figure~\ref{fig:framework}), which addresses the limitations of zero-shot ESD generation.
\texttt{SIF} is a two-staged framework which fine-tunes LM on a small set of ESDs in the first stage. In the second stage, ESDs generated using the fine-tuned LM are passed through a sequence of Ro{BERT}a-based classifiers~\cite{liu2019roberta} to identify relevant events, remove redundant events, and predict pair-wise temporal ordering between the events.
These pair-wise orderings are then used to create a full event ordering using topological sorting on a directed graph created from the predicted orderings.
\begin{table}[t!]
\centering
\small
\begin{tabular}{p{7.3cm}}
\hline 
\textbf{\textsc{Sequence}} here is a sequence of events that happen while baking a cake: $1$. $e_1$ $2$. $e_2$\\
\textbf{\textsc{Expect}} these are the things that happen when you bake a cake: $1$. $e_1$ $2$. $e_2$\\
\textbf{\textsc{Ordered}} here is an ordered sequence of events that occur when you bake a cake: $1$. $e_1$ $2$. $e_2$\\
\textbf{\textsc{Describe}} describe baking a cake in small sequences of short sentences: $1$. $e_1$ $2$. $e_2$\\
\textbf{\textsc{Direct}} baking a cake: $1$. $e_1$ $2$. $e_2$\\
\textbf{\textsc{Tokens}} $\langle$\texttt{SCR}$\rangle$ baking a cake $\langle$\texttt{ESCR}$\rangle$: $1$. $e_1$ $2$. $e_2$\\
\textbf{\textsc{AllTokens}} $\langle$\texttt{SCR}$\rangle$ baking a cake $\langle$\texttt{ESCR}$\rangle$: $\langle$\texttt{BEVENT}$\rangle$ $e_1$ $\langle$\texttt{EEVENT}$\rangle$ $\langle$\texttt{BEVENT}$\rangle$ $e_2$ $\langle$\texttt{EEVENT}$\rangle$\\
\hline
\end{tabular}
\caption{\label{tab:prompts} Different prompt formulations for \textsc{Baking a cake} scenario with two events ($e_1$ and $e_2$). }
\end{table}
\subsection{Stage I: Fine-tuning PLMs}
PLMs fine-tuned on commonsense datasets like ATOMIC~\citep{sap2019atomic} can generalize beyond the scenarios observed during fine-tuning~\citep{bosselut-etal-2019-comet}.
Hence, we investigate the learning capability of LMs when a small number of script examples are available. We fine-tune LMs on ESDs using different natural language and pseudo-natural language prompt formulations for encoding ESDs (Table~\ref{tab:prompts}) to study the effect of prompt formulations on this task as observed during the probing experiments. We fine-tune LMs using 
negative log-likelihood objective. 
\subsection{Stage II: Post-processing Generated ESDs}
We sample ESDs for an unseen scenario using the fine-tuned LMs and employ a $3$-step post-processing method to correct them for relevance, repetitions, and ordering.
\subsubsection{Step 1: Irrelevant Events Removal}
The first post-processing step is to remove non-scenario-relevant events from an ESD. An event is not relevant for a scenario if it is not a part of the scenario (e.g., `tipping a waiter' is not a part of \textsc{baking a cake} scenario). For irrelevant events removal, we first need to identify irrelevant events for a scenario. We pose this identification problem as a binary classification task to predict if a given event belongs to a given scenario. For training purpose, a positive example is constructed by pairing a scenario with an event belonging to that scenario; negative samples are drawn from another scenario in the training data. Using this data, we train a Ro{BERT}a-L-based~\cite{liu2019roberta} classifier and remove those events from an ESD which are predicted as irrelevant by this classifier.
\subsubsection{Step 2: Event De-duplication}
The second step involves the identification and removal of repeated events. Repetition of events can occur by an exact copy of an event or by a paraphrase of an event (e.g., `6. You have your batter prepared` and `7. You have your batter made' in $5$ of Table~\ref{tab:probing}). To identify such de-duplications, we train a RoBERTa-L-based paraphrase identification system using MRPC~\citep{dolan2005automatically} dataset. However, we observe many false-positives (e.g., `open a faucet' and `close a faucet' were identified as paraphrases) with this system. Since false-positives can lead to unnecessary removal of events, we employ a conservative approach of only identifying repeated events. We find edit distance between each pair of events in an ESD and remove multiple occurrences of an event from the ESD, as identified by the edit distance score of $0$. 
\subsubsection{Step 3: Temporal Order Correction}
The final step is to correct the order of events in an ESD. We correct the ESDs for ordering by first obtaining pair-wise event orderings and then using a graph-based approach to get the final overall ordering. We pose the problem of pair-wise event ordering as a binary classification task to predict if the order of a given pair of events is correct with respect to the given scenario. We sample event pairs from gold ESDs to construct positive (sequence order) and negative (reverse order) examples to train a Ro{BERT}a-L-based classifier.
Topological sort is then used to get the final ESD for a scenario from the ordering predictions for all the $N\choose2$ pairs of events in an ESD. We construct a directed graph $\mathcal{G}=(\mathcal{V},\mathcal{E})$ of events in a scenario with events as nodes ($\mathcal{V}$) of the graph and a directed edge from node $v_1 \in \mathcal{V}$ to $v_2 \in \mathcal{V}$ if event represented by $v_2 $ is predicted to occur after the event represented by $v_1$. We keep the original ordering of events in case the constructed graph is cyclic\footnote{$66\pm15$\% (averaged across all the input variants and folds) of the complete graphs are acyclic for GPT$2$.} due to incorrect predictions from the classifiers. 

\subsection{Implementation Details}
\subsubsection{Dataset pre-proccessing}
We fine-tune LMs on ESDs from DeScript~\citep{wanzare2016crowdsourced} dataset which consists of $100$ ESDs each for $40$ scenarios, collected via crowdsourcing. The scenarios are randomly partitioned into $8$ folds with each fold consisting of ESDs from $5$ scenarios to perform $8$-fold cross-validation of \texttt{SIF} for each of the prompt formulation. We lowercase and enclose each ESD within a begin of scenario \texttt{$\langle$BOS$\rangle$} and an end of scenario \texttt{$\langle$EOS$\rangle$} token for fine-tuning. The input to the relevance classifier is: \texttt{scenario} \texttt{$\langle$/s$\rangle$} \texttt{$e$} and to the temporal classifier is \texttt{scenario name} \texttt{$\langle$/s$\rangle$} \texttt{$e_1$} \texttt{$\langle$/s$\rangle$} \texttt{$e_2$}, where \texttt{$\langle$/s$\rangle$} is a separator token and $e$, $e_1$, $e_2$ are events.
\subsubsection{Training details}
We use huggingface's transformers library~\citep{wolf-etal-2020-transformers} to fine-tune LMs on each of the $7$ prompt formulations, leading to $7$ variations for each LM, for $1$ epoch with a batch size of $1$, gradient accumulation per $16$ steps, and block size of $150$. At inference time, $5$ ESDs are sampled for each of the given scenarios with top $50$ probable tokens, nucleus sampling~\citep{holtzman2019curious} probability of $0.9$, and maximum length set at $150$. We use Ro{BERT}a-L architecture from the transformers library for relevance and temporal order classifiers. Relevance (Temporal) classifier is trained for $10$ ($5$) epochs with average validation accuracy of $84.50\%$ ($83.87\%$) across the folds.  The model with the best accuracy on the valid split is used in the post-processing stage. We use python's \textit{editdistance} library to compute edit distance for the de-duplication step.
We use Adam optimizer with an initial learning rate of $2e^{-5}$, warm-up steps set at $0.06$ of total steps, batch size of $16$, and maximum input length $150$ for both the classifiers. All the models are trained and tested on NVIDIA Tesla V$100$ SXM$2$ $16$GB GPU machine.

\begin{table*}[t!]
\centering
\scriptsize
\begin{tabular}{l|ccccccc}
\textbf{Models} & \textbf{\textsc{Tokens}} & \textbf{\textsc{Expect}} &	\textbf{\textsc{Sequence}} &	\textbf{\textsc{AllTokens}} &	\textbf{\textsc{Describe}} &\textbf{\textsc{Direct}} &	\textbf{\textsc{Ordered}} \\
\hline
(1) Zero-shot & $03.1$ $(5.2)$&	 $03.6$ $(5.5)$&	$05.4$ $(2.8)$ &	 $03.1$ $(5.2)$&	$03.2$ $(3.6)$ &	 $03.9$ $(5.1)$&	$06.2$ $(6.6)$\\
(2) GPT$2$-L$_{\textsc{SCRATCH}}$ & $17.2$ $(3.1)$&	 $19.3$ $(3.7)$&	$16.8$ $(2.9)$ &	 $18.6$ $(4.5)$&	$17.6$ $(2.6)$ &	 $14.4$ $(3.9)$&	$17.7$ $(3.2)$\\
\hline
(3) {BART}-FT & $15.5$ $(6.0)$&	 $20.8$ $(3.5)$&	$19.6$ $(3.5)$ &	 $19.7$ $(9.2)$&	$19.2$ $(3.9)$ &	 $18.0$ $(6.6)$&	$11.7$ $(4.8)$\\
(4) {GPT$2$}-FT & $30.7$ $(5.1)$&	 $31.3$ $(5.5)$&	$32.4$ $(6.3)$ &	 $30.7$ $(6.6)$&	$32.3$ $(5.9)$ &	 $31.4$ $(5.8)$&	$31.0$ $(4.8)$\\
\hline
(5) {BART}-\textbf{\texttt{SIF}} & $16.8$ $(5.1)$ & $21.1$ $(4.2)$&	$19.9$ $(3.7)$ &	$20.5$ $(11.1)$&	 $20.0$ $(3.8)$&	 $19.6$ $(7.2)$&	$13.7$  $(5.0)$\\
(6) GPT$2$-\textbf{\texttt{SIF}} & $\mathbf{33.6}$ $(5.4)$ & $\mathbf{33.9}$ $(5.6)$&	$\mathbf{35.2}$ $(6.9)$ &	$\mathbf{32.5}$ $(6.9)$&	 $\mathbf{34.2}$ $(5.3)$&	 $\mathbf{33.6}$ $(5.7)$&	$\mathbf{33.2}$  $(5.5)$\\
\hline
\end{tabular}
\caption{\label{tab:automatic} Automatic evaluation results: Mean BLEU scores (and std. dev.) over $8$ folds of held-out scenarios are reported. (1) is pre-trained GPT$2$ (no fine-tuning or post-processing); (2) is randomly initialized GPT$2$ with fine-tuning; (3-4) are fine-tuned BART and GPT$2$; (5-6) are \textbf{\texttt{SIF}} applied to BART and GPT$2$.}
\end{table*}
\begin{table*}[t!]
\centering
\resizebox{\textwidth}{!}{
\begin{tabular}{l|ccccccc}
\textbf{Models} & \textbf{\textsc{Tokens}} & \textbf{\textsc{Expect}} &	\textbf{\textsc{Sequence}} &	\textbf{\textsc{AllTokens}} &	\textbf{\textsc{Describe}} &\textbf{\textsc{Direct}} &	\textbf{\textsc{Ordered}} \\
\hline
(1) {GPT$2$}-FT & $30.7$ $(5.1)$&	 $31.3$ $(5.5)$&	$32.4$ $(6.3)$ &	 $30.7$ $(6.6)$&	$32.3$ $(5.9)$ &	 $31.4$ $(5.8)$&	$31.0$ $(4.8)$\\
(2) {GPT$2$}-FT+Relevance (R) &  $33.1$ $(5.1)$&	 $33.1$ $(4.9)$ 	&$34.7$ $(6.9)$&	$31.9$ $(6.7)$&	 $33.7$ $(5.0)$	&$32.6$ $(5.8)$&	$33.2$ $(5.2)$\\
(3) {GPT$2$}-FT+R+De-duplicate (D) & $33.5$ $(5.2)$&	 $33.6$ $(5.2)$&	 $35.1$ $(6.9)$&	$32.1$ $(6.7)$&	 $\mathbf{34.3}$ $(5.0)$&	 $32.9$ $(5.7)$&	$\mathbf{33.6}$ $(5.5)$\\
\hline
(4) {GPT$2$}-FT+R+D+Reorder (GPT$2$-\texttt{SIF}) & $\mathbf{33.6}$ $(5.4)$ & $\mathbf{33.9}$ $(5.6)$&	$\mathbf{35.2}$ $(6.9)$ &	$\mathbf{32.5}$ $(6.9)$&	 $34.2$ $(5.3)$&	 $\mathbf{33.6}$ $(5.7)$&	$33.2$  $(5.5)$\\
 \hline
 \end{tabular}
 }
\caption{\label{tab:ablation} Ablation analysis of each step in the proposed pipeline for GPT$2$. Mean BLEU scores (and std. dev.) over $8$ folds of held-out scenarios are reported. (1) fine-tuned GPT$2$; (2-4) are fine-tuned GPT$2$ with successive post-processing steps.}
\end{table*}
\section{Evaluation}
We use \texttt{SIF} to induce script knowledge in GPT$2$, BART, and T$5$, and evaluate full ESDs generated for a given unseen scenario using \textbf{BLEU} metric~\citep{papineni-etal-2002-bleu}, following~\citet{pichotta2016using} who use BLEU to score individual LM-generated events. As BLEU is a precision-based metric, we measure n-gram overlap of the sampled ESDs against multiple gold-reference ESDs\footnote{We use NLTK python library to calculate BLEU score with add-$1$ smoothing function and n-grams upto $n=4$. We convert the outputs of different variants \& gold references into numbered form, $1$. $e_1$ $2$. $e_2$ \dots $n$. $e_n$ for a fair comparison.} for each scenario in the test fold. 

Additionally, for deeper analysis of the generated ESDs, two of the authors evaluate a subset of the generated ESDs (blinded to the identity of the models and prompt variants) on three levels -- individual events (\textbf{Relevance (R)}), pairwise events (\textbf{Order (O)}), and the overall sequence (\textbf{Missing (M)}). \textbf{R} measures the \% of generated events relevant to a scenario; \textbf{O} measures the \% of consecutive event pairs correctly ordered given a scenario; and \textbf{M} measures the degree to which important events are missing on a $4$-point Likert scale defined as ($1$) no or almost no missing events, ($2$) some insignificant missing events, ($3$) notable missing events, and ($4$) severe missing events. As scripts are complex structures and require an understanding of scenarios, we chose not to resort to a crowdsourcing platform for manual analysis. We manually analyze the outputs to evaluate \texttt{SIF} as well as perform an error analysis to identify opportunities for future research directions.

We evaluate our framework on scenarios in each of the eight folds as well as novel scenarios from \citet{regneri-etal-2010-learning}, and day-to-day activities. As we do not have access to gold-reference ESDs for the novel scenarios, we demonstrate our framework's performance only using manual evaluation.

\section{Results and Analysis}\label{sec:results}

\subsection{Automatic Evaluation}
We present the automatic evaluation results on held-out scenarios 
in Table~\ref{tab:automatic}. As baselines, we report scores from non-fine-tuned GPT$2$-L (Zero-shot), a randomly-initialized GPT$2$-L$_\textsc{SCRATCH}$ model fine-tuned on DeScript ESDs, and BART-FT and GPT$2$-FT models which are fine-tuned in the first stage of \texttt{SIF}. We do not report any results for T$5$ as it was even struggling to learn the input ESD formulations during fine-tuning. We explain the findings from automatic evaluation below.

\noindent \textbf{\texttt{SIF} significantly outperforms fine-tuning baselines.}~Both GPT$2$-\texttt{SIF} and BART-\texttt{SIF} have higher BLEU scores as compared to their corresponding fine-tuned (GPT$2$-FT and BART-FT) models across all the prompt variants. This clearly reflects the advantage of the post-processing stage in \texttt{SIF} framework. Improvement across different LMs reinforces the LM-agnostic nature of our framework. Variation in the extent of induction across prompt variants indicates the sensitivity of LMs to prompt formulations. 

\noindent \textbf{Script knowledge is best accessible through GPT$\mathbf{2}$ than other LMs.}~As previously mentioned in probing experiments, BART and T$5$ were not able to output anything useful in the zero-shot setting while GPT$2$ could produce ESDs, although erroneous and of poor quality. We observe same trends even after fine-tuning these LMs or using \texttt{SIF} to induce script knowledge in these LMs. Interestingly, a randomly initialized and fine-tuned GPT$2$ (GPT$2$-L$_\textsc{SCRATCH}$) is able to perform comparable to a pre-trained BART fined-tuned using DeScript (BART-FT), and even better for \textsc{Tokens} and \textsc{Ordered} variants. 
Overall, GPT$2$ is found to be better than BART in terms of the presence and accessibility of script knowledge through them. One possible explanation for this is that GPT$2$ is a generative language model while BART and T$5$ are encoder-decoder-based language models making it challenging to encode complete script knowledge within a scenario name. 

\noindent \textbf{Performance across LMs is sensitive to prompt formulation and scenario.}~We consistently observe variation in performance across prompt variants. Moreover, this variation is also observed across LMs. For BART, \textsc{Expect} outperforms other prompt variants while \textsc{Sequence} performs the best for GPT$2$. High variance across folds also shows that different prompts perform differently depending upon a scenario. This indicates the sensitivity of LMs to prompt formulations and thus justifies our experiments with different prompt formulations to study the extent of script knowledge that can be accessed through PLMs.
\begin{table}[t!]
\centering
\resizebox{\columnwidth}{!}{
\begin{tabular}{l|c|ccc}
\multirow{2}{*}{\textbf{Variants}} & \multirow{2}{*}{\textbf{BLEU}$\uparrow$} &   \multicolumn{3}{c}{\textbf{Manual Evaluation}} \\
 &&\textbf{R}$\uparrow$ & \textbf{O}$\uparrow$ & \textbf{M}$\downarrow$\\
\hline 
\textsc{Tokens} & $19.2/\mathbf{22.8}$ & $77.2/\mathbf{84.3}$ & $72.3/\mathbf{89.3}$ & $\mathbf{2.6}/\mathbf{2.6}$\\
\textsc{Expect} & $22.8/\mathbf{26.0}$  & $81.9/\mathbf{82.7}$& $74.5/\mathbf{86.5}$ & $\mathbf{3.0}/\mathbf{3.0}$\\
\textsc{Sequence} & $27.8/\mathbf{33.4}$  &$73.3/\mathbf{83.2}$ & $74.0/\mathbf{87.5}$  & $\underline{\mathbf{2.5}}/\underline{\mathbf{2.5}}$\\
\textsc{AllTokens} & $\underline{33.5}/\underline{\mathbf{35.0}}$& $83.5/\mathbf{85.7}$&$82.7/\underline{\mathbf{89.5}}$& $\mathbf{2.6}/\mathbf{2.6}$\\
\textsc{Describe} & $27.1/\mathbf{28.6}$  & $80.7/\underline{\mathbf{86.3}}$& $83.9/\mathbf{85.9}$& $\mathbf{2.8}/\mathbf{2.8}$\\ 
\textsc{Direct} & $30.9/\mathbf{34.1}$  & $81.2/\mathbf{84.2}$& $\underline{\mathbf{88.5}}/86.1$ & $\mathbf{2.6}/\mathbf{2.6}$\\
\textsc{Ordered} & $\mathbf{31.9}/31.5$ & $\underline{84.9}/\mathbf{86.2}$& $78.6/\mathbf{86.8}$& $\mathbf{2.6}/\mathbf{2.6}$ \\
\hline
\end{tabular}
}
\caption{ 
Manual and BLEU scores on fine-tuned GPT$2$ (GPT$2$-FT) \texttt{SIF} applied to GPT$2$ (FT/\texttt{SIF}), computed for a stratified sample of outputs (one ESD per scenario across two folds). Mean scores across two annotators are reported.
Annotator agreement is measured with Cohen's Kappa~\citep{cohen1960coefficient} ($\kappa$=$0.61$ for \textbf{O}, $\kappa$=$0.56$ for \textbf{R})
and Spearman's correlation ($\rho$=$0.64$ for \textbf{M}). \underline{Underline} and \textbf{bold} denotes the best across variants, and between FT and Ours, respectively. O scores are calculated only when both the events are marked as relevant by the two annotators.} 
\label{tab:automate-manual}
\end{table}
\subsection{Ablation Analysis of \textbf{\texttt{SIF}}}
We next analyze the contribution of each the stage of \texttt{SIF} and each step of stage~II leading to improvement in the performance via an ablation study, on GPT$2$, in Table~\ref{tab:ablation}. As expected stage~I contributes maximum to the performance boost. and
There is a consistent improvement in BLEU after each of the post-processing steps except in the case of \textsc{Describe} and \textsc{Ordered} wherein, reordering leads to a slight decrease in BLEU as the trained classifiers are not perfectly accurate. We present qualitative outputs when \texttt{SIF} is used to induce script knowledge in GPT$2$ in Table~\ref{tab:held-out}.

\subsection{Manual Evaluation and Error Analysis}
We manually evaluate a total of $140$ ESDs (for M) comprising $652$ individual events (for R) and $582$ consecutive pair of events (for O) generated from GPT$2$-FT and GPT$2$~\texttt{SIF} across all the prompt variants (Table~\ref{tab:automate-manual}). 
BLEU scores are also reported for the same set of ESDs to study the correlation between manual and automatic metrics.
We find that outputs from \texttt{SIF} have higher BLEU, R, and O scores than FT across all prompt variants (except O for \textsc{Direct} and BLEU for \textsc{Ordered}). M scores do not change, which shows that significantly important events are not dropped during the irrelevant events removal step. Different prompts perform well in different aspects. \textsc{Describe} generates most relevant events, \textsc{AllTokens} has the best temporal ordering knowledge, and \textsc{Sequence} leads to least severe missing events after Stage II of \texttt{SIF}.
To our surprise, we find no statistically significant correlation between BLEU and any of the manual evaluation metrics (pearson correlation between BLEU and R, O and M was $r=0.23,-0.06,-0.49$ with p$>0.1$, respectively), emphasizing a need for more sophisticated automatic metrics than BLEU for evaluating full ESDs, having a complex structure. The best performing variant as per BLEU score differs from the best one in Table~\ref{tab:automatic} due to variance in performance across scenarios as well as different sampled ESDs of the same scenario in Table~\ref{tab:automate-manual}.
\begin{table}[t!]
\centering
\scriptsize
\resizebox{\columnwidth}{!}{
\begin{tabular}{lrrr}
\hline 
\textbf{Scenario} & \textbf{R}$\uparrow$ & \textbf{O}$\uparrow$ & \textbf{M}$\downarrow$\\
\hline 
Order fastfood online &$81.5$ &$84.6$ &$2.6$\\ 
Cook in a microwave &$89.5$ & $92.0$& $2.4$\\
Answer telephone  &$65.5$ &$91.7$ &$2.0$\\
Buy from vending machine &$77.1$ & $81.3$&$3.4$ \\
Tie shoe laces  & $65.8$ & $66.7$&$3.6$\\
Brush teeth& $75.9$& $71.4$&$2.6$\\
Make ginger paste& $41.5$& $85.7$&$3.4$\\
Attend a wedding& $71.9$& $100.0$&$2.4$\\
Wash a car & $85.7$&$90.0$&$3.0$\\
Take out trash& $88.5$&$92.3$ &$2.2$\\
Take a taxi &$85.7$ &$76.2$ & $2.0$\\
Surf the internet &$73.3$ & $62.5$&$2.8$ \\
Watch television &$77.4$ & $73.7$& $3.0$\\
Go to a club to dance &$100.0$ & $93.5$& $1.4$\\
\hline
Average Score & $77.1$ & $83.0$ & $2.6$\\
\hline
\end{tabular}
}
\caption{\label{tab:manual} Manual evaluation of ESDs for novel scenarios. Averaged across $5$ sampled ESDs per scenario generated using the best performing \textsc{Sequence} variant of GPT$2$-\texttt{SIF} as per automatic measure.}
\end{table}
\begin{table*}[t!]
\centering
\scriptsize
{\begin{tabular}{p{1.5cm}p{0.3cm}p{13cm}}
\hline
&& \hspace{4cm} \textbf{Good quality generations}\\
\hline
\multirow{3}{*}{\parbox{1.5cm}{\centering \textsc{going on a train}}} & FT & 1. get dressed 2. go to station 3. buy ticket 4. get on train 5. wait for train 6. get on train 7. sit in seat 8. read newspaper 9. wait for train 10. get off train 11. get dressed 12. leave station\\
& \texttt{SIF} &  1. go to station 2. buy ticket 3. wait for train 4. get on train 5. sit in seat 6. get off train 7. leave station\\
\hline
\multirow{3}{*}{\parbox{1.5cm}{\centering \textsc{flying in an airplane}}}& FT & 1. go to airport 2. get boarding pass 3. go to the airline counter 4. present boarding pass and boarding pass id to attendant 5. present ticket and id to attendant 6. sit down in seat 7. make sure you are in seat 8. board airplane\\
& \texttt{SIF} & 1. go to airport 2. go to the airline counter 3. present ticket and id to attendant 4. present boarding pass and boarding pass id to attendant 5. get boarding pass 6. board airplane 7. make sure you are in seat 8. sit down in seat \\
\hline
&& \hspace{4cm} \textbf{Poor quality generations}\\
\hline
\multirow{6}{*}{\parbox{1.5cm}{\centering \textsc{cleaning up a flat}}} & FT & 1. get the flat 2. take the flat to the bathroom 3. take some water 4. take the flat to the sink 5. turn on the water 6. let it wash the flat 7. then put in some dryer sheet 8. get a sponge 9. scrub the flat 10. put the flat in the dryer 11. when done dry the flat with a towel 12. put away the flat\\
& \texttt{SIF} & 1. get the flat 2. take the flat to the bathroom 3. take some water 4. take the flat to the sink 5. turn on the water 6. let it wash the flat 7. then put in some dryer sheet 8. get a sponge 9. scrub the flat 10. put the flat in the dryer 11. when done dry the flat with a towel 12. put away the flat\\
\hline
\multirow{3}{*}{\parbox{1.5cm}{\centering \textsc{taking the underground}}} & FT & 1. arrive at the designated location 2. get in car 3. park in car 4. walk into the designated area 5. enter the room 6. find a chair 7. sit on a chair 8. watch movie \\
& \texttt{SIF} & 1. walk into the designated area \\
\hline
\end{tabular}}
\caption{\label{tab:held-out} Scripts generated using \textsc{Sequence} variant of GPT$2$ for held-out scenarios. FT denotes output from the fine-tuned model and \texttt{SIF} refers to outputs from our framework applied to GPT$2$.}
\end{table*}
\begin{table*}[t!]
\centering
\scriptsize
{\begin{tabular}{p{1.5cm}p{0.3cm}p{13cm}}
\hline
&& \hspace{4cm} \textbf{Good quality generations}\\
\hline
\multirow{2}{*}{\parbox{1.5cm}{\centering \textsc{Brushing teeth}}} & FT & 1. go into bathroom 2. turn on faucet 3. remove any dirt or debris 4. grab a brush 5. scrub and floss the teeth 6. leave the bathroom\\
& \texttt{SIF} & 1. go into bathroom 2. grab a brush 3. scrub and floss the teeth 4. leave the bathroom\\
\hline
\multirow{3}{*}{\parbox{1.5cm}{\centering \textsc{Going to a club to dance}}}& FT & 1. choose which club to attend. 2. drive or park your car. 3. get in your car. 4. go to the club. 5. enter the club. 6. get up and dance.\\
& \texttt{SIF} & 1. choose which club to attend. 2. get in your car. 3. go to the club. 4. drive or park your car. 5. enter the club. 6. get up and dance.\\
\\
\hline
\multirow{2}{*}{\parbox{1.5cm}{\centering \textsc{taking a taxi}}} & FT & 1. get in car 2. get into car 3. wait for taxi 4. enter the car 5. pay the fare 6. get out the driver 7. get out the door 8. exit car\\
& \texttt{SIF} & 1. get into car 2. get in car 3. wait for taxi 4. enter the car 5. pay the fare 6. get out the driver 7. exit car 8. get out the door\\
\hline
&& \hspace{4cm} \textbf{Poor quality generations}\\
\hline
\multirow{6}{*}{\parbox{1.5cm}{\centering \textsc{Making ginger paste}}} & FT &  1. get your hot water 2. get your bowl 3. turn on the hot water 4. whisk a bowl of sugar into a paste 5. put the bowl on the stove 6. turn on the hot water 7. boil the paste 8. add salt to the paste 9. turn off the water 10. put the bowl on a rack 11. pour the hot water into a saucepan 12. put some salt and sugar in the saucepan 13. turn the heat on 14. pour the sauce on to the bowl 15. eat the paste\\
& \texttt{SIF} & 1. get your hot water 2. get your bowl 3. turn on the hot water 4. whisk a bowl of sugar into a paste 5. put the bowl on the stove 6. boil the paste 7. add salt to the paste 8. put the bowl on a rack 9. pour the hot water into a saucepan 10. put some salt and sugar in the saucepan 11. turn the heat on 12. pour the sauce on to the bowl 13. eat the paste\\
\hline
\multirow{3}{*}{\parbox{1.5cm}{\centering \textsc{surfing the internet}}} & FT & 1. get in the car. 2. drive to the beach. 3. rent a surfboard. 4. go to the beach. 5. put on my sunglasses. 6. get in my car. 7. drive to my beach. 8. go to the surfing.\\
& \texttt{SIF} & 1. 1. rent a surfboard. 2. get in the car. 3. go to the surfing.\\
\hline
\multirow{7}{*}{\parbox{1.5cm}{\centering \textsc{attending a wedding}}} & FT & 1. schedule an appointment for the reception 2. drive to the venue 3. show your guest information 4. present your wedding ceremony/ reception to the guests 5. invite the friends/ family/ coworkers/ etc. to the wedding 7. present your wedding present to the guests 8. make the guest list 9. take the invitation 10. go to the reception 11. pay the fee/ gratuity 12. give your guest his/her ticket 13. leave the venue\\
& \texttt{SIF} & 1. schedule an appointment for the reception 2. drive to the venue 3. show your guest information 4. present your wedding ceremony/ reception to the guests 5. invite the friends/ family/ coworkers/ etc. to the wedding 6. present your wedding present to the guests 7. make the guest list 8. take the invitation 9. go to the reception 10. leave the venue\\
\hline
\multirow{4}{*}{\parbox{1.5cm}{\centering \textsc{buying from a vending machine}}} & FT &
1. enter the shop 2. look for the item you wish to purchase 3. go into the register to purchase the item 4. pay for the item 5. take your item to your car 6. take the item to the car and remove the card 7. leave the shop 8. return the card 9. leave the shop\\
& \texttt{SIF} & 1. enter the shop 2. look for the item you wish to purchase 3. go into the register to purchase the item 4. pay for the item 5. take your item to your car 6. take the item to the car and remove the card 7. leave the shop 8. return the card\\
\hline
\end{tabular}}
\caption{\label{tab:novel-output} Scripts generated using \textsc{Sequence} variant of GPT$2$ for novel scenarios. FT denotes output from the fine-tuned model and \texttt{SIF} refers to outputs from our framework applied to GPT$2$\protect \footnotemark.}
\end{table*}

Manual evaluation revealed that a model can miss significant events, even though it can generate many relevant ones. As we only de-duplicate multiple occurrences of exactly the same events in a scenario, we observe repeated paraphrases ($4.6\%$ across all prompt variants) of the same event, such as `pour some milk in the pot' and `pour the milk into the coffee pot' (\textsc{making coffee} scenario). $23.9\%$ of the irrelevant events ($13.5\%$ across all prompt variants) are incoherent (`take the flat to the bathroom' for \textsc{cleaning a flat}),  $11.4\%$ mixed (`sit in front of coffee shop' for \textsc{making coffee}), $61.4\%$ unrelated (`add shampoo' for \textsc{washing dishes}), and rest ungrammatical.

We present a manual evaluation of novel scenarios to gauge the generalizability of our framework in Table~\ref{tab:manual}.  The framework generalizes to most of the novel scenarios except for those which involve very granular events like \textsc{making ginger paste} or \textsc{tying shoe laces}. Although GPT$2$ is a contextualized model, it confuses \textsc{buying from vending machine} with buying from a store, \textsc{surfing the internet} with the `surfing' activity, or \textsc{attending a wedding} with `getting married'. Additionally, we provide a few good and bad quality outputs from GPT$2$ models for held-out (Table~\ref{tab:held-out}) and novel (Table~\ref{tab:novel-output}) scenarios to identify the avenues for improving script induction in LMs.
\section{Limitations}
\noindent \textbf{De-duplication of Events.}~As mentioned previously, \texttt{SIF} cannot de-deuplicate paraphrased version of an event. Therefore, more sophisticated paraphrase identification systems could be used to de-duplicate such events. There could be scenarios where multiple occurrence of same event is required. For instance, \textsc{washing dishes} wherein faucet needs to be opened and closed once at the starting before applying soap and secondly after applying soap (when washed by hands). Hence, it is required to differentiate between desirable and undesirable repetition of events.

\noindent \textbf{Full vs Partial Temporal Ordering.}~While we consider the task of generating full event sequence descriptions for a scenario, we acknowledge that many scenarios may not have strict ordering of events (e.g., either wet ingredients can be mixed first or dry ones in a \textsc{baking a cake} scenario) or there can be overlapping events (e.g., while oven is pre-heating, batter can be prepared). Instead of considering partial ordering of events~\citep{sakaguchi2021proscript}, we focus on generating multiple possible full sequence of events for a scenario and report the averaged scores.

\section{Conclusion and Future Work}
We investigate whether pre-trained language models are capable of generating full event sequence descriptions with minimal prompting and find that pre-trained GPT$2$ has an incomplete understanding of scripts, while BART and T$5$ did not even produce anything useful through zero-shot probing experiments. We propose \texttt{SIF}, an LM-agnostic script induction framework, that is shown to produce meaningful ESDs for unseen scenarios and mitigate errors (such as scenario-irrelevant, repeated, and misordered events) that were observed during probing experiments, as measured by automatic and manual evaluation. We also provide evidence for the generalization capability of our framework to novel scenarios. However, there is great room for improvement which is evident from manual error analysis and qualitative outputs. Future work may focus on developing more sophisticated automatic metrics as well as an end-to-end system for script induction which might help in mitigating cascading of errors, due to each component, common to any pipeline-based approaches.

\section*{Acknowledgements}
We would like to thank Benjamin Van Durme, Sweta Agrawal, and all the anonymous reviewers for their
valuable feedback ad suggestions.
\bibliography{anthology,custom}

\begin{thebibliography}{41}
\expandafter\ifx\csname natexlab\endcsname\relax\def\natexlab#1{#1}\fi

\bibitem[{Bosselut et~al.(2019)Bosselut, Rashkin, Sap, Malaviya, Celikyilmaz,
  and Choi}]{bosselut-etal-2019-comet}
Antoine Bosselut, Hannah Rashkin, Maarten Sap, Chaitanya Malaviya, Asli
  Celikyilmaz, and Yejin Choi. 2019.
\newblock \href {https://doi.org/10.18653/v1/P19-1470} {{COMET}: Commonsense
  transformers for automatic knowledge graph construction}.
\newblock In \emph{Proceedings of the 57th Annual Meeting of the Association
  for Computational Linguistics}, pages 4762--4779, Florence, Italy.
  Association for Computational Linguistics.

\bibitem[{Bouraoui et~al.(2020)Bouraoui, Camacho-Collados, and
  Schockaert}]{bouraoui2020inducing}
Zied Bouraoui, Jose Camacho-Collados, and Steven Schockaert. 2020.
\newblock Inducing relational knowledge from {BERT}.
\newblock In \emph{Proceedings of the AAAI Conference on Artificial
  Intelligence}, volume~34, pages 7456--7463.

\bibitem[{Chambers(2017)}]{chambers2017behind}
Nathanael Chambers. 2017.
\newblock Behind the scenes of an evolving event cloze test.
\newblock In \emph{Proceedings of the 2nd Workshop on Linking Models of
  Lexical, Sentential and Discourse-level Semantics}, pages 41--45.

\bibitem[{Chambers and Jurafsky(2008)}]{chambers2008unsupervised}
Nathanael Chambers and Dan Jurafsky. 2008.
\newblock Unsupervised learning of narrative event chains.
\newblock In \emph{Proceedings of ACL-08: HLT}, pages 789--797.

\bibitem[{Chambers and Jurafsky(2009)}]{chambers2009unsupervised}
Nathanael Chambers and Dan Jurafsky. 2009.
\newblock Unsupervised learning of narrative schemas and their participants.
\newblock In \emph{Proceedings of the Joint Conference of the 47th Annual
  Meeting of the ACL and the 4th International Joint Conference on Natural
  Language Processing of the AFNLP}, pages 602--610.

\bibitem[{Cohen(1960)}]{cohen1960coefficient}
Jacob Cohen. 1960.
\newblock A coefficient of agreement for nominal scales.
\newblock \emph{Educational and psychological measurement}, 20(1):37--46.

\bibitem[{Devlin et~al.(2018)Devlin, Chang, Lee, and
  Toutanova}]{devlin2018bert}
Jacob Devlin, Ming-Wei Chang, Kenton Lee, and Kristina Toutanova. 2018.
\newblock {BERT}: Pre-training of deep bidirectional transformers for language
  understanding.
\newblock \emph{arXiv preprint arXiv:1810.04805}.

\bibitem[{Dolan and Brockett(2005)}]{dolan2005automatically}
William~B Dolan and Chris Brockett. 2005.
\newblock Automatically constructing a corpus of sentential paraphrases.
\newblock In \emph{Proceedings of the Third International Workshop on
  Paraphrasing (IWP2005)}.

\bibitem[{Feldman et~al.(2019)Feldman, Davison, and
  Rush}]{feldman2019commonsense}
Joshua Feldman, Joe Davison, and Alexander~M Rush. 2019.
\newblock Commonsense knowledge mining from pretrained models.
\newblock \emph{arXiv preprint arXiv:1909.00505}.

\bibitem[{Gordon and Van~Durme(2013)}]{gordon2013reporting}
Jonathan Gordon and Benjamin Van~Durme. 2013.
\newblock Reporting bias and knowledge acquisition.
\newblock In \emph{Proceedings of the 2013 workshop on Automated knowledge base
  construction}, pages 25--30.

\bibitem[{Holtzman et~al.(2019)Holtzman, Buys, Du, Forbes, and
  Choi}]{holtzman2019curious}
Ari Holtzman, Jan Buys, Li~Du, Maxwell Forbes, and Yejin Choi. 2019.
\newblock The curious case of neural text degeneration.
\newblock \emph{arXiv preprint arXiv:1904.09751}.

\bibitem[{Lin et~al.(2020)Lin, Lee, Khanna, and Ren}]{lin2020birds}
Bill~Yuchen Lin, Seyeon Lee, Rahul Khanna, and Xiang Ren. 2020.
\newblock Birds have four legs?! numersense: Probing numerical commonsense
  knowledge of pre-trained language models.
\newblock \emph{arXiv preprint arXiv:2005.00683}.

\bibitem[{Liu et~al.(2019)Liu, Ott, Goyal, Du, Joshi, Chen, Levy, Lewis,
  Zettlemoyer, and Stoyanov}]{liu2019roberta}
Yinhan Liu, Myle Ott, Naman Goyal, Jingfei Du, Mandar Joshi, Danqi Chen, Omer
  Levy, Mike Lewis, Luke Zettlemoyer, and Veselin Stoyanov. 2019.
\newblock Ro{BERT}a: A {R}obustly {O}ptimized {BERT} {P}retraining {A}pproach.
\newblock \emph{arXiv preprint arXiv:1907.11692}.

\bibitem[{Lyu et~al.(2020)Lyu, Zhang, and
  Callison-Burch}]{lyu-zhang-wikihow:2020}
Qing Lyu, Li~Zhang, and Chris Callison-Burch. 2020.
\newblock \href
  {http://www.cis.upenn.edu/~ccb/publications/reasoning-about-goals-with-wikihow.pdf}
  {Reasoning about goals, steps, and temporal ordering with wikihow}.
\newblock In \emph{Proceedings of The 2020 Conference on Empirical Methods In
  Natural Language Proceedings (EMNLP)}.

\bibitem[{Miikkulainen(1995)}]{miikkulainen1995script}
Risto Miikkulainen. 1995.
\newblock Script-based inference and memory retrieval in subsymbolic story
  processing.
\newblock \emph{Applied Intelligence}, 5(2):137--163.

\bibitem[{Modi et~al.(2017)Modi, Anikina, Ostermann, and
  Pinkal}]{modi2017inscript}
Ashutosh Modi, Tatjana Anikina, Simon Ostermann, and Manfred Pinkal. 2017.
\newblock Inscript: Narrative texts annotated with script information.
\newblock \emph{arXiv preprint arXiv:1703.05260}.

\bibitem[{Modi and Titov(2014)}]{modi2014inducing}
Ashutosh Modi and Ivan Titov. 2014.
\newblock Inducing neural models of script knowledge.
\newblock In \emph{Proceedings of the Eighteenth Conference on Computational
  Natural Language Learning}, pages 49--57.

\bibitem[{Mostafazadeh et~al.(2016)Mostafazadeh, Chambers, He, Parikh, Batra,
  Vanderwende, Kohli, and Allen}]{mostafazadeh2016corpus}
Nasrin Mostafazadeh, Nathanael Chambers, Xiaodong He, Devi Parikh, Dhruv Batra,
  Lucy Vanderwende, Pushmeet Kohli, and James Allen. 2016.
\newblock A corpus and cloze evaluation for deeper understanding of commonsense
  stories.
\newblock In \emph{Proceedings of the 2016 Conference of the North American
  Chapter of the Association for Computational Linguistics: Human Language
  Technologies}, pages 839--849.

\bibitem[{Mueller(2004)}]{mueller2004understanding}
Erik~T Mueller. 2004.
\newblock Understanding script-based stories using commonsense reasoning.
\newblock \emph{Cognitive Systems Research}, 5(4):307--340.

\bibitem[{Ostermann(2020)}]{ostermann2020script}
Simon Ostermann. 2020.
\newblock Script knowledge for natural language understanding.

\bibitem[{Ostermann et~al.(2018)Ostermann, Modi, Roth, Thater, and
  Pinkal}]{ostermann2018mcscript}
Simon Ostermann, Ashutosh Modi, Michael Roth, Stefan Thater, and Manfred
  Pinkal. 2018.
\newblock Mcscript: A novel dataset for assessing machine comprehension using
  script knowledge.
\newblock \emph{arXiv preprint arXiv:1803.05223}.

\bibitem[{Ostermann et~al.(2019)Ostermann, Roth, and
  Pinkal}]{ostermann2019mcscript2}
Simon Ostermann, Michael Roth, and Manfred Pinkal. 2019.
\newblock Mcscript2. 0: A machine comprehension corpus focused on script events
  and participants.
\newblock \emph{arXiv preprint arXiv:1905.09531}.

\bibitem[{Papineni et~al.(2002)Papineni, Roukos, Ward, and
  Zhu}]{papineni-etal-2002-bleu}
Kishore Papineni, Salim Roukos, Todd Ward, and Wei-Jing Zhu. 2002.
\newblock \href {https://doi.org/10.3115/1073083.1073135} {{B}leu: a method for
  automatic evaluation of machine translation}.
\newblock In \emph{Proceedings of the 40th Annual Meeting of the Association
  for Computational Linguistics}, pages 311--318, Philadelphia, Pennsylvania,
  USA. Association for Computational Linguistics.

\bibitem[{Petroni et~al.(2020)Petroni, Lewis, Piktus, Rockt{\"a}schel, Wu,
  Miller, and Riedel}]{petroni2020context}
Fabio Petroni, Patrick Lewis, Aleksandra Piktus, Tim Rockt{\"a}schel, Yuxiang
  Wu, Alexander~H Miller, and Sebastian Riedel. 2020.
\newblock How context affects language models' factual predictions.
\newblock \emph{arXiv preprint arXiv:2005.04611}.

\bibitem[{Pichotta and Mooney(2016)}]{pichotta2016using}
Karl Pichotta and Raymond~J Mooney. 2016.
\newblock Using sentence-level lstm language models for script inference.
\newblock \emph{arXiv preprint arXiv:1604.02993}.

\bibitem[{Radford et~al.(2019)Radford, Wu, Child, Luan, Amodei, and
  Sutskever}]{radford2019language}
Alec Radford, Jeffrey Wu, Rewon Child, David Luan, Dario Amodei, and Ilya
  Sutskever. 2019.
\newblock Language models are unsupervised multitask learners.
\newblock \emph{OpenAI blog}, 1(8):9.

\bibitem[{Regneri et~al.(2010)Regneri, Koller, and
  Pinkal}]{regneri-etal-2010-learning}
Michaela Regneri, Alexander Koller, and Manfred Pinkal. 2010.
\newblock \href {https://www.aclweb.org/anthology/P10-1100} {Learning script
  knowledge with web experiments}.
\newblock In \emph{Proceedings of the 48th Annual Meeting of the Association
  for Computational Linguistics}, pages 979--988, Uppsala, Sweden. Association
  for Computational Linguistics.

\bibitem[{Rudinger et~al.(2015)Rudinger, Rastogi, Ferraro, and
  Van~Durme}]{rudinger-etal-2015-script}
Rachel Rudinger, Pushpendre Rastogi, Francis Ferraro, and Benjamin Van~Durme.
  2015.
\newblock \href {https://doi.org/10.18653/v1/D15-1195} {Script induction as
  language modeling}.
\newblock In \emph{Proceedings of the 2015 Conference on Empirical Methods in
  Natural Language Processing}, pages 1681--1686, Lisbon, Portugal. Association
  for Computational Linguistics.

\bibitem[{Sakaguchi et~al.(2021)Sakaguchi, Bhagavatula, Bras, Tandon, Clark,
  and Choi}]{sakaguchi2021proscript}
Keisuke Sakaguchi, Chandra Bhagavatula, Ronan~Le Bras, Niket Tandon, Peter
  Clark, and Yejin Choi. 2021.
\newblock proscript: Partially ordered scripts generation via pre-trained
  language models.
\newblock \emph{arXiv preprint arXiv:2104.08251}.

\bibitem[{Sap et~al.(2019)Sap, Le~Bras, Allaway, Bhagavatula, Lourie, Rashkin,
  Roof, Smith, and Choi}]{sap2019atomic}
Maarten Sap, Ronan Le~Bras, Emily Allaway, Chandra Bhagavatula, Nicholas
  Lourie, Hannah Rashkin, Brendan Roof, Noah~A Smith, and Yejin Choi. 2019.
\newblock Atomic: An atlas of machine commonsense for if-then reasoning.
\newblock In \emph{Proceedings of the AAAI Conference on Artificial
  Intelligence}, volume~33, pages 3027--3035.

\bibitem[{Schank and Abelson(1975)}]{schank1975scripts}
Roger~C Schank and Robert~P Abelson. 1975.
\newblock Scripts, plans, and knowledge.
\newblock In \emph{IJCAI}, volume~75, pages 151--157.

\bibitem[{Shwartz et~al.(2020)Shwartz, West, Bras, Bhagavatula, and
  Choi}]{shwartz2020unsupervised}
Vered Shwartz, Peter West, Ronan~Le Bras, Chandra Bhagavatula, and Yejin Choi.
  2020.
\newblock Unsupervised commonsense question answering with self-talk.
\newblock \emph{arXiv preprint arXiv:2004.05483}.

\bibitem[{Singh et~al.(2002)Singh, Lin, Mueller, Lim, Perkins, and
  Zhu}]{singh2002open}
Push Singh, Thomas Lin, Erik~T Mueller, Grace Lim, Travell Perkins, and Wan~Li
  Zhu. 2002.
\newblock Open mind common sense: Knowledge acquisition from the general
  public.
\newblock In \emph{OTM Confederated International Conferences" On the Move to
  Meaningful Internet Systems"}, pages 1223--1237. Springer.

\bibitem[{Wanzare et~al.(2017{\natexlab{a}})Wanzare, Zarcone, Thater, and
  Pinkal}]{wanzare2017inducing}
Lilian Wanzare, Alessandra Zarcone, Stefan Thater, and Manfred Pinkal.
  2017{\natexlab{a}}.
\newblock Inducing script structure from crowdsourced event descriptions via
  semi-supervised clustering.

\bibitem[{Wanzare et~al.(2017{\natexlab{b}})Wanzare, Zarcone, Thater, and
  Pinkal}]{wanzare-etal-2017-inducing}
Lilian Wanzare, Alessandra Zarcone, Stefan Thater, and Manfred Pinkal.
  2017{\natexlab{b}}.
\newblock \href {https://doi.org/10.18653/v1/W17-0901} {Inducing script
  structure from crowdsourced event descriptions via semi-supervised
  clustering}.
\newblock In \emph{Proceedings of the 2nd Workshop on Linking Models of
  Lexical, Sentential and Discourse-level Semantics}, pages 1--11, Valencia,
  Spain. Association for Computational Linguistics.

\bibitem[{Wanzare et~al.(2016)Wanzare, Zarcone, Thater, and
  Pinkal}]{wanzare2016crowdsourced}
Lilian~DA Wanzare, Alessandra Zarcone, Stefan Thater, and Manfred Pinkal. 2016.
\newblock A crowdsourced database of event sequence descriptions for the
  acquisition of high-quality script knowledge.

\bibitem[{Weir et~al.(2020)Weir, Poliak, and Van~Durme}]{weir2020probing}
Nathaniel Weir, Adam Poliak, and Benjamin Van~Durme. 2020.
\newblock Probing neural language models for human tacit assumptions.
\newblock CogSci.

\bibitem[{Wolf et~al.(2020)Wolf, Debut, Sanh, Chaumond, Delangue, Moi, Cistac,
  Rault, Louf, Funtowicz, Davison, Shleifer, von Platen, Ma, Jernite, Plu, Xu,
  Le~Scao, Gugger, Drame, Lhoest, and Rush}]{wolf-etal-2020-transformers}
Thomas Wolf, Lysandre Debut, Victor Sanh, Julien Chaumond, Clement Delangue,
  Anthony Moi, Pierric Cistac, Tim Rault, Remi Louf, Morgan Funtowicz, Joe
  Davison, Sam Shleifer, Patrick von Platen, Clara Ma, Yacine Jernite, Julien
  Plu, Canwen Xu, Teven Le~Scao, Sylvain Gugger, Mariama Drame, Quentin Lhoest,
  and Alexander Rush. 2020.
\newblock \href {https://doi.org/10.18653/v1/2020.emnlp-demos.6} {Transformers:
  State-of-the-art natural language processing}.
\newblock In \emph{Proceedings of the 2020 Conference on Empirical Methods in
  Natural Language Processing: System Demonstrations}, pages 38--45, Online.
  Association for Computational Linguistics.

\bibitem[{Zhang et~al.(2020)Zhang, Chen, Wang, Song, and
  Roth}]{zhang2020analogous}
Hongming Zhang, Muhao Chen, Haoyu Wang, Yangqiu Song, and Dan Roth. 2020.
\newblock Analogous process structure induction for sub-event sequence
  prediction.
\newblock \emph{arXiv preprint arXiv:2010.08525}.

\bibitem[{Zhou et~al.(2020)Zhou, Ning, Khashabi, and Roth}]{zhou2020temporal}
Ben Zhou, Qiang Ning, Daniel Khashabi, and Dan Roth. 2020.
\newblock Temporal common sense acquisition with minimal supervision.
\newblock \emph{arXiv preprint arXiv:2005.04304}.

\bibitem[{Zhou et~al.(2019)Zhou, Shah, and Schockaert}]{zhou2019learning}
Yilun Zhou, Julie~A Shah, and Steven Schockaert. 2019.
\newblock Learning household task knowledge from wikihow descriptions.
\newblock \emph{arXiv preprint arXiv:1909.06414}.

\end{thebibliography}

\appendix

\section{Appendix}
\label{sec:appendix}
\subsection{Experimental Details}\label{sec:exp-details}
Train time for fine-tuning GPT$2$-L ($774$M parameters) on each of the folds for each of the variant was $13(\pm1)$ minutes on an average. Ro{BERT}a-L architecture has $355$M parameters and it took $282$ minutes on an average to train relevance classifier for each of the folds while $294$ minutes for temporal ordering classifier. Relevance and temporal ordering classifiers were trained on each of the training fold by keeping aside all the ESDs from a randomly chosen scenario for validation purposes as shown in Table~\ref{tab:fold}.
\begin{table*}[h]
\centering
\scriptsize
\resizebox{0.85\textwidth}{!}
{\begin{tabular}{p{0.6cm}|p{13cm}|p{3cm}}
\hline 
\textbf{Fold} & \hspace{4cm}\textbf{Scenarios} & \hspace{1cm}\textbf{Held-out} \\
\hline 
$1$ & baking a cake, borrowing a book from the library, flying in an airplane, going on a train, riding on a bus & cooking pasta\\
$2$ & getting a hair cut, going grocery shopping, planting a tree, repairing a flat bicycle tire, taking a bath & going bowling\\
$3$ & eating in a fast food restaurant, paying with a credit card, playing tennis, going to the theater, taking a child to bed & planting a tree\\
$4$ & washing dishes, making a bonfire, going to the sauna, making coffee, going to the swimming pool & going grocery shopping\\
$5$ & taking a shower, ironing laundry, taking a driving lesson, going to the dentist, going to a funeral & taking the underground\\
$6$ & washing one's hair, fueling a car, sending food back (in a restaurant), changing batteries in an alarm clock, checking in at an airport & paying with a credit card\\
$7$ & having a barbecue, ordering a pizza, cleaning up a flat, making scrambled eggs, taking the underground & eating in a fast food restaurant\\
$8$ & renovating a room, cooking pasta, sewing a button, doing laundry, going bowling & getting a hair cut\\
\hline
\end{tabular}}
\caption{\label{tab:fold} Dataset partitioning details. Held-out refers to the scenarios kept aside from training-folds for validation purposes of relevance and temporal ordering classifiers.}
\end{table*}
\subsection{Qualitative Outputs}
We present a few outputs from BART-FT and BART-\texttt{SIF} in Table~\ref{tab:bart-output}. We observe that the outputs are of poor quality. Although \texttt{SIF} is able to filter out a few irrelevant events in all the examples and correct the ordering (e.g. \textsc{going on a train}), there are many mistakes like the events are not coherent `board the train Borders Boarders', or repetitions such as `enjoy the train', `enjoy train'. One reason for poor performance is that BART is an encode-decoder-based language model and thus encoding script knowledge in a scenario is a challenging task. On the other hand, GPT$2$ is a decoder-based language model which is trained with a language modeling objective and is able to generate the following events given a scenario in the form of a natural language prompt.
\begin{table*}[b!] 
\centering
\scriptsize
\resizebox{0.85\textwidth}{!}
{\begin{tabular}{p{1.5cm}p{0.3cm}p{13cm}}
\hline
\multirow{8}{*}{\parbox{1.5cm}{\centering \textsc{going on a train}}} & FT & 1 2. go to the train station 3. board 4. get off 5. board the train Borders Borders 7. get a ticket 8. find a seat 9. enjoy train Borders 10. get dressed 11. ride the train 12. get home barriers 13. enjoy the train 14. pay for travel Alpha Alpha Omega Omega Omega Alpha Omega Delta Alpha Omega Alpha Alpha Alpha Delta Omega Beta Delta Delta Gamma Delta Delta Delta DDDDFDFEE bundles baggage transfer WTBSD and luggage transfer 14. get on board 15. get luggage assigned to correct location Bold TGBS Framework 6. ride to terminal DWDSD\\
& \texttt{SIF} &  1. go to the train station 2. get a ticket 3. get luggage assigned to correct location  Bold TGBS  Framework 4. board the train  Borders  Borders 5. get on board 6. board 7. ride to terminal  DWDSD  TGBS 8. find a seat 9. enjoy the train 10. ride the train 11. enjoy train  Borders 12. get off 13. get home  barriers\\
\hline
\multirow{5}{*}{\parbox{1.5cm}{\centering \textsc{having a barbeque}}} & FT & 1. set up the grill 2. put some water in the water 3. bring the grill out 4. throw a large amount of charcoal in the pit 5. open the fire 6. throw the lighter fluid into the fire 7. add the newspaper 8. wait for the flame to get hot 9. put a small amount of grease on the wood 10. enjoy it\\
& \texttt{SIF} & 1. set up the grill 2. bring the grill out 3. throw a large amount of charcoal in the pit 4. open the fire 5. throw the lighter fluid into the fire 6. wait for the flame to get hot 7. put a small amount of grease on the wood 8. enjoy it\\
\hline
\multirow{7}{*}{\parbox{1.5cm}{\centering \textsc{borrowing a book from a library}}}& FT & 1. call the library 2. find a book to borrow 3. sign the book release paper 4. obtain the book 5. place the book in the book shelf 6. read the book reaching your full potential 7. go home autoritarian reaching the goal 8. pay for the book 9. start the car the goal 10. continue until you reach it 11. put the book away 12. check the cover of the book 13. read your favorite book 14. put a receipt in your wallet demonstrating that you read the correct book aut autoritarian style of aut aut\\
& \texttt{SIF} & 1. call the library 2. find a book to borrow 3. obtain the book 4. pay for the book 5. check the cover of the book 6. read your favorite book 7. continue until you reach it 8. sign the book release paper 9. place the book in the book shelf 10. read the book  reaching your full potential 11. put the book away\\
\hline
\end{tabular}}
\caption{\label{tab:bart-output} Scripts generated using \textsc{Expect} variant of BART for held-out scenarios. FT denotes output from the fine-tuned model and \texttt{SIF} refers to outputs from our framework when applied to BART. We filter extra tokens (\texttt{$\langle$s$\rangle$}, \texttt{$\langle$/s$\rangle$}, \texttt{$\langle$SEP$\rangle$}) generated in between the events to present a clean output. These extra tokens were not generated in the case of GPT$2$.}
\end{table*}
\end{document}